\documentclass[a4paper, 10pt, conference]{llncs}      

\usepackage{graphicx} 
\usepackage{amssymb}  
\usepackage[noend]{algpseudocode}
\usepackage{subfigure}
\usepackage{dblfloatfix} 
\usepackage{makecell}
\usepackage{hyperref}
\hypersetup{
    colorlinks=true,
}
\usepackage[linesnumbered, lined, ruled]{algorithm2e}
\usepackage{textcmds}
\usepackage{acro}

\DeclareAcronym{ad}{
	short=AD,
	long=Autonomous Driving,
}

\DeclareAcronym{adas}{
	short=ADAS,
	long=Advance Driver Assistance Systems,
}
\DeclareAcronym{hd}{
	short=HD,
	long=High Definition,
}
\DeclareAcronym{gnn}{
	short=GNN,
	long=Graph Neural Network,
}
\DeclareAcronym{bev}{
	short=BEV,
	long=Bird-Eye-View,
}
\DeclareAcronym{msda}{
	short=MSDA,
	long=Multi-Scale-Deformable-Attention,
}
\DeclareAcronym{cnn}{
	short=CNN,
	long=Convolution Neural Networks,
}
\DeclareAcronym{detr}{
	short=DETR,
	long=Detection Transformer,
}
\DeclareAcronym{dino}{
	short=DINO,
	long=DETR with Improved deNoising anchOr box,
}
\DeclareAcronym{caspformer}{
	short=CASPFormer,
	long=Context Aware Scene Prediction Transformer,
}
\DeclareAcronym{caspnet}{
	short=CASPNet,
	long=Context Aware Scene Prediction Network,
}
\DeclareAcronym{mlp}{
	short=MLP,
	long=Multi-Layer Perceptron,
}

\title{\LARGE \bf
CASPFormer: Trajectory Prediction from BEV Images with Deformable Attention
}

\author{Harsh Yadav\inst{1} \and Maximilian Sch\"afer\inst{2} \and Kun Zhao\inst{2} \and Tobias Meisen\inst{1}}
\institute{University of Wuppertal, Germany \email{\{harsh.yadav, meisen\}@uni-wuppertal.de}\and Aptiv Services Deutschland GmbH \email{\{maximilian.schaefer, kun.zhao\}@aptiv.com}}

\begin{document}

\maketitle

\begin{abstract}
Motion prediction is an important aspect for \ac{ad} and \ac{adas}. Current state-of-the-art motion prediction methods rely on \ac{hd} maps for capturing the surrounding context of the ego vehicle. Such systems lack scalability in real-world deployment as \ac{hd} maps are expensive to produce and update in real-time. To overcome this issue, we propose \ac{caspformer}, which can perform multi-modal motion prediction from rasterized \ac{bev} images. Our system can be integrated with any upstream perception module that is capable of generating \ac{bev} images. Moreover, \ac{caspformer} directly decodes vectorized trajectories without any post-processing. Trajectories are decoded recurrently using deformable attention, as it is computationally efficient and provides the network with the ability to focus its attention on the important spatial locations of the \ac{bev} images. In addition, we also address the issue of mode collapse for generating multiple scene-consistent trajectories by incorporating learnable mode queries. We evaluate our model on the nuScenes dataset and show that it reaches state-of-the-art across multiple metrics.
\end{abstract}


\section{Introduction}\label{section:introduction}
In recent years, \ac{ad} and \ac{adas} technologies have gained huge attention as they can significantly improve the safety and comfort standards across the automotive industry \cite{liang2020learning}. The current approach to these self-driving tasks is to divide them into multiple independent sub-tasks, mainly i) perception, ii) motion prediction, and iii) motion planning, and optimize each task individually \cite{chen2023end}. The perception task deals with the detection and segmentation of surrounding dynamic and static environment contexts. The dynamic context captures the motion of the dynamic agents in the scene e.g. pedestrians, cyclists, vehicles, traffic lights, etc., while the static context includes stationary elements of the scene e.g. road and lane boundaries, pedestrian crossings, traffic signs, parked vehicles, construction sites etc. As defined by  Cui et al. \cite{cui2019multimodal}, the motion prediction task involves predicting multi-modal future trajectories for agents in a scene. The prediction of multiple future trajectories enables the model to account for uncertainties in the dynamic context. In addition, to ensure safety critical operation, the predicted trajectories must adhere to the static and dynamic contexts. Lastly, the objective of the motion planning task is to generate the control actions for the ego vehicle to navigate it through the scene while adhering to the traffic rules and dynamics of the vehicle.


\begin{figure}[t]
\centering
\includegraphics[width=122mm]{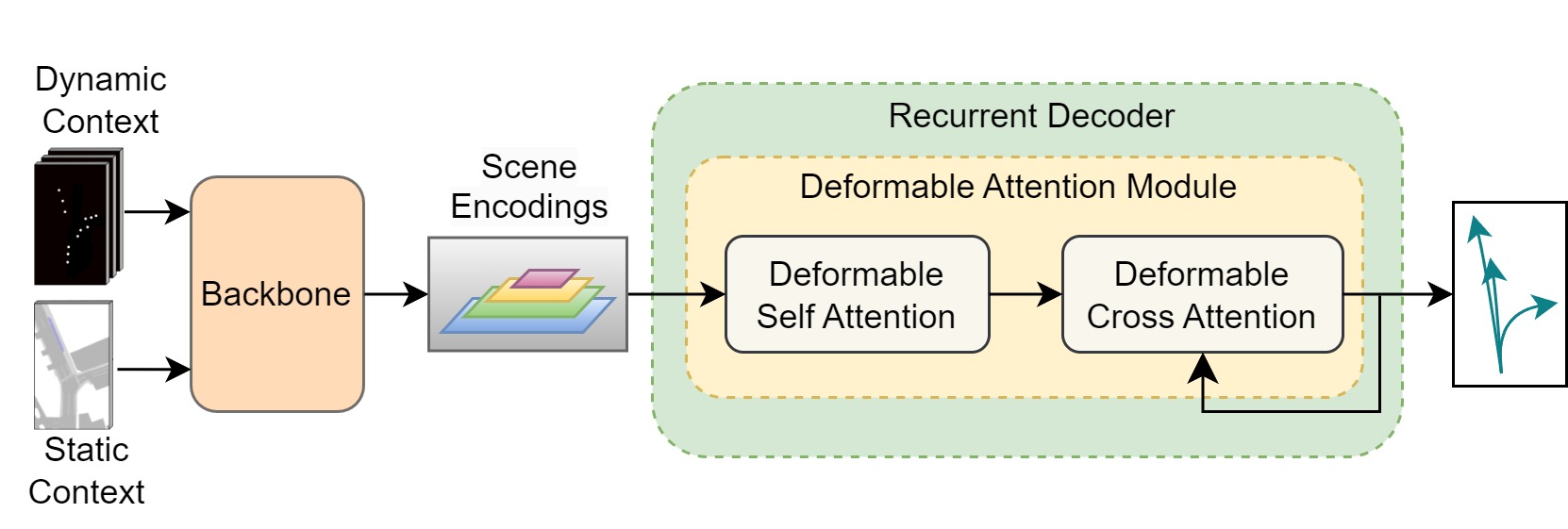}
\caption{Shows an overview of the CASPFormer architecture. The backbone uses CNN and convolution RNN to generate the scene encodings. The scene encodings have a pyramid structure with increasing resolution from top to bottom. The deformable self-attention module applies a multi-scale feature fusion on the scene encodings, while the deformable cross-attention module recurrently decodes the trajectories. The output of the previous time step is used to update the position of the reference point and the query embeddings in deformable cross-attention.}
\label{fig:overall-architecture}
\end{figure}

Current state-of-the-art models \cite{ngiam2021scene,girgis2021latent,schafer2022context,zhou2022hivt,luo2023jfp,zhou2023query} in motion prediction require \ac{hd} maps for static context with centimeter-level accuracy. Such a strict constraint on \ac{hd} maps leads to high production costs \cite{casas2021mp3}. Thus, these models suffer from the problem of scalability in a real-world deployment. A cost-effective and scalable alternative is to construct \ac{bev} images from a vision perception system deployed on the ego vehicle, as proposed by Li et al. \cite{li2022bevformer} in their BEVFormer model. To efficiently decode trajectories and learn spatial attention on the feature maps of \ac{bev} images, we opt for the deformable attention mechanism proposed in Deformable \ac{detr} \cite{zhu2021deformable}. Furthermore, to generate a diverse set of modes in multi-modal trajectory prediction, we incorporate learnable embeddings into our architecture. Contrary to previous studies \cite{girgis2021latent,varadarajan2022multipath++,zhou2023query}, which use one set of learnable embeddings, our network consists of two sets of learnable embeddings. The first set, temporal queries, is responsible for capturing the temporal correlation in the output trajectories, and the second set, mode queries, aims to address the issue of mode collapse. Following the works \cite{schafer2022context,zhou2023query}, we recurrently decode the multi-modal trajectories. This allows the network to update the reference point for deformable attention and the temporal queries through feedback loops of the recurrent decoder. 


A depiction of our proposed network \ac{caspformer} is shown in Figure \ref{fig:overall-architecture}. Furthermore, Figure \ref{fig:recurrent-decoder} highlights the components of the recurrent decoder. The contributions of our work are summarised as follows:
\begin{itemize}
    \item A novel motion prediction architecture is introduced that generates multi-modal vectorized trajectories from \ac{bev} images.
    \item It incorporates two sets of learnable embeddings: temporal queries for capturing the temporal correlation in the output trajectories and mode queries for overcoming the issue of mode collapse.
    \item The trajectory decoding is done recurrently using deformable attention where the feedback loops update the reference point for deformable attention and the temporal queries. 
    \item We evaluate our method on the nuScenes motion prediction benchmark \cite{nuscene2020prediction} and show that it achieves state-of-the-art performance across various metrics.
\end{itemize}

\section{Related Work}
In this section, we highlight the corresponding related work. Section \ref{section:scene representation} categorized the previous studies based on how their scene representation is constructed. Section \ref{section:multi-modal-prediction} highlights various methods for generating multi-modal prediction. Section \ref{section: transformer based attention} discusses several transformer-based attention mechanisms that can be used to extract meaningful representations from \ac{bev} images.

\subsection{Input Scene Representation}
\label{section:scene representation}
The scene representation in the motion prediction task can be divided into two categories, rasterized scene representation and vectorized scene representation. The studies with rasterized scene representation \cite{cui2019multimodal,chai2019multipath,gilles2021home,schafer2022context} take advantage of matured practices in \ac{cnn} to extract scene encodings. On the other hand, the vectorized representation was first introduced by LaneGCN \cite{liang2020learning} which identified that \ac{hd} maps have an underlying graph structure that can be exploited to learn long-range and efficient static scene encodings with \ac{gnn}. VectorNet \cite{gao2020vectornet} later showed that not only the static context, but also the dynamic context can also be represented in vectorized format. Follow-up studies \cite{girgis2021latent,zhou2022hivt,varadarajan2022multipath++,zhou2023query} have provided several motion prediction methods that receive both static and dynamic contexts in vectorized form. 




\subsection{Multi-Modal Prediction}\label{section:multi-modal-prediction}
To accommodate uncertainties in traffic scenarios, autonomous vehicles must predict various scene-consistent trajectories adhering to the static and dynamic context. One approach \cite{casas2020implicit,cui2021lookout} employs a variational auto-encoder to learn multiple latent representations of the entire scene and then decodes these latent representations generating multiple trajectories corresponding to each agent. However, these methods require multiple forward passes during both training and inference and are prone to mode collapse. Other approaches \cite{gilles2021home,schafer2022context} use spatial-temporal grids to predict the future position for each agent and sample multiple goal positions. Thereafter, scene-consistent trajectories are generated which connect the proposed goal positions with the current position of the agents. These approaches learn multi-modality inherently without a specific training strategy, however, post-processing is required to generate trajectories from the grid. Alternatively, Multipath \cite{chai2019multipath} utilizes fixed anchors corresponding to different modes. It constructs multiple trajectories by generating the offsets and probability distribution corresponding to each one of these anchors. A potential limitation of Multipath is that most of the fixed anchors are not relevant for particular scenes. This issue is addressed in the follow-up studies \cite{girgis2021latent,varadarajan2022multipath++,zhou2023query}, which learn the anchors during the training with the help of learnable embeddings and predict a diverse set of modes.

\subsection{Transformer-based Attention in Image Domain}
\label{section: transformer based attention}
In recent years, transformer-based attention \cite{vaswani2017attention} mechanisms have achieved huge success in the image domain. The studies \cite{dosovitskiy2021an,liu2021swin,xia2022vision} establish the foundation for transformer-based encoders for image processing. Since these approaches lack decoder networks, their application is limited only to feature extraction. On the contrary, \ac{detr} \cite{carion2020end} introduces a transformer-based encoder-decoder architecture capable of end-to-end object detection. However, \ac{detr} suffers from two major problems: slow convergence and low performance in detecting small objects, as its encoder is limited to processing features with very small resolution due to its quadratic computational complexity with the size of feature maps.

Deformable \ac{detr} \cite{zhu2021deformable} overcomes these problems by sparsifying the selection of values and computing the attention solely based upon queries whilst eliminating the need for keys. The decrease in computational cost allows both the encoder and decoder to attend to every feature map in the feature pyramid generated by the backbone. Deformable \ac{detr} thus significantly reduces training time while increasing performance in detecting small objects. Follow-up studies \cite{roh2021sparse,li2023lite} on Deformable \ac{detr} establish that a large part of its computational cost comes from the deformable self-attention module, and therefore propose to reduce this cost by limiting the number of queries which undergo self-attention. We compare training time with and without deformable self-attention modules in ablation studies because computational cost plays an important role in the deployment of models on edge devices operating in vehicles. 

\section{Methods}\label{section:methods}
This section will explain the methods which are utilized in our work and in particular our contribution to the current state of the art. Section \ref{section:input_output_formulation} describes the formulation of the input and output of the network. Section \ref{section:network_architecture} focuses on network architecture of \ac{caspformer} and its components. Section \ref{section:loss_formulation} illustrates the loss formulation.

\subsection{Input-Output Formulation}
\label{section:input_output_formulation}
\ac{caspformer} receives static and dynamic contexts of the surrounding region of the ego vehicle and outputs multi-modal vectorized trajectories. 

\noindent \textbf{Static Context Input.} The static context is rasterized into a grid-based input of shape $(H, W)$. The feature dimension of rasterized static context contains binary feature maps consisting of information about the derivable area, center lines, driving lanes, road boundaries, and pedestrian crossing. The input of static context can be depicted as follows:
\begin{equation}
    I_{s} \in \mathbb{R}^{H \times W \times \mid F_s\mid},
\end{equation}
where $H$ is the height of the grid, $W$ is the width of the grid, and $\mid F_s \mid$ is the number of input features of static context. 



\noindent \textbf{Dynamic Context Input.} The dynamic context contains the motion information of all the surrounding road agents for the past $T_i$ time steps. Corresponding to each time
step $T_i$, a grid of shape $(H, W)$ is created. The feature dimension of these grids contains the velocity, acceleration, location offset, height, width, and heading information. The rasterized input of dynamic context is as follows:
\begin{equation}
    I_{d} \in \mathbb{R}^{T_i \times H \times W \times \mid F_d\mid},
\end{equation}
where, $\mid F_d \mid$ is the number of input features of dynamic context. 

\noindent \textbf{Output.} The predicted trajectories contain the position information i.e. $(x,y)$ of the ego vehicle, and the output tensor can thus be represented as:
\begin{equation}
    Y \in \mathbb{R}^{M\times T_o \times 2},
\end{equation}

\noindent where, $M$ is the number of modes, $T_o$ is the number of future time steps.

\begin{figure}
        \centering
	\includegraphics[width=122mm]{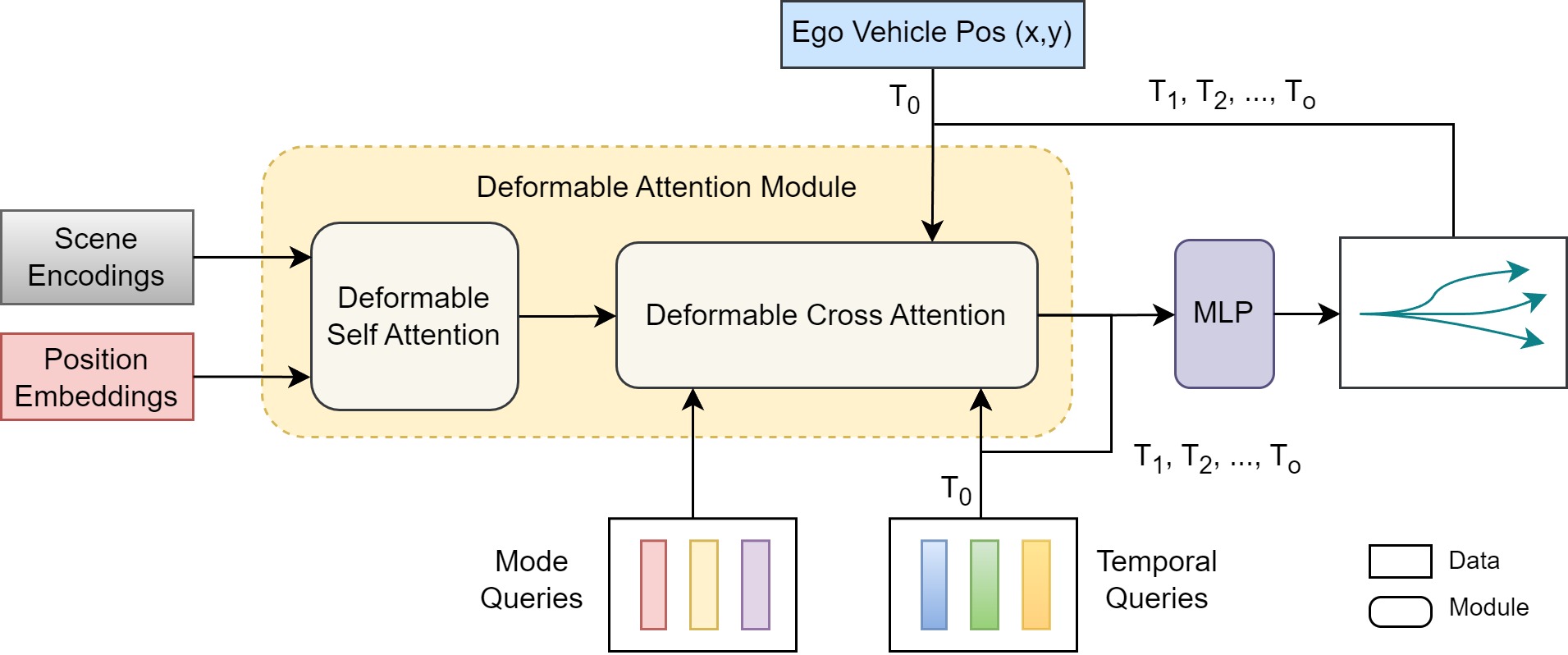}
        \caption{A depiction of the recurrent decoder network architecture. The position embeddings are non-learnable and help the network in learning the location of features. The mode queries serve the purpose of producing multiple scene-consistent trajectories in the multi-modal output. The temporal correlation in the predicted trajectories is captured with temporal queries. The position of the reference point for the deformable attention is set to the ego vehicle position in the scene. The recurrent architecture updates the ego vehicle position and the temporal queries at every recurrent step.}
 \label{fig:recurrent-decoder}
\end{figure}

\subsection{Network Architecture}
\label{section:network_architecture}
The overall network architecture is shown in Figure \ref{fig:overall-architecture}. The network consists of a backbone and a recurrent decoder. For our work, the backbone architecture is adopted from \ac{caspnet} \cite{schafer2022context}, as it is currently state-of-the-art in the nuScenes dataset \cite{caesar2020nuscenes}. It receives static and dynamic contexts in rasterized formats to generate multi-scale scene encodings. It is important to note that the \ac{caspformer} is not limited to a particular backbone and can be extended to other transformer or \ac{cnn} based backbones. The works \cite{schafer2022context,zhou2023query} suggests that decoding the trajectory in a recurrent fashion results in better prediction capabilities. Inspired by this observation, we also decode the trajectory recurrently from the multi-scale scene encodings. A detailed schematic of the recurrent decoder is depicted in Figure \ref{fig:recurrent-decoder}.

\begin{figure}[t]
    \centering
    \includegraphics[width=77mm]{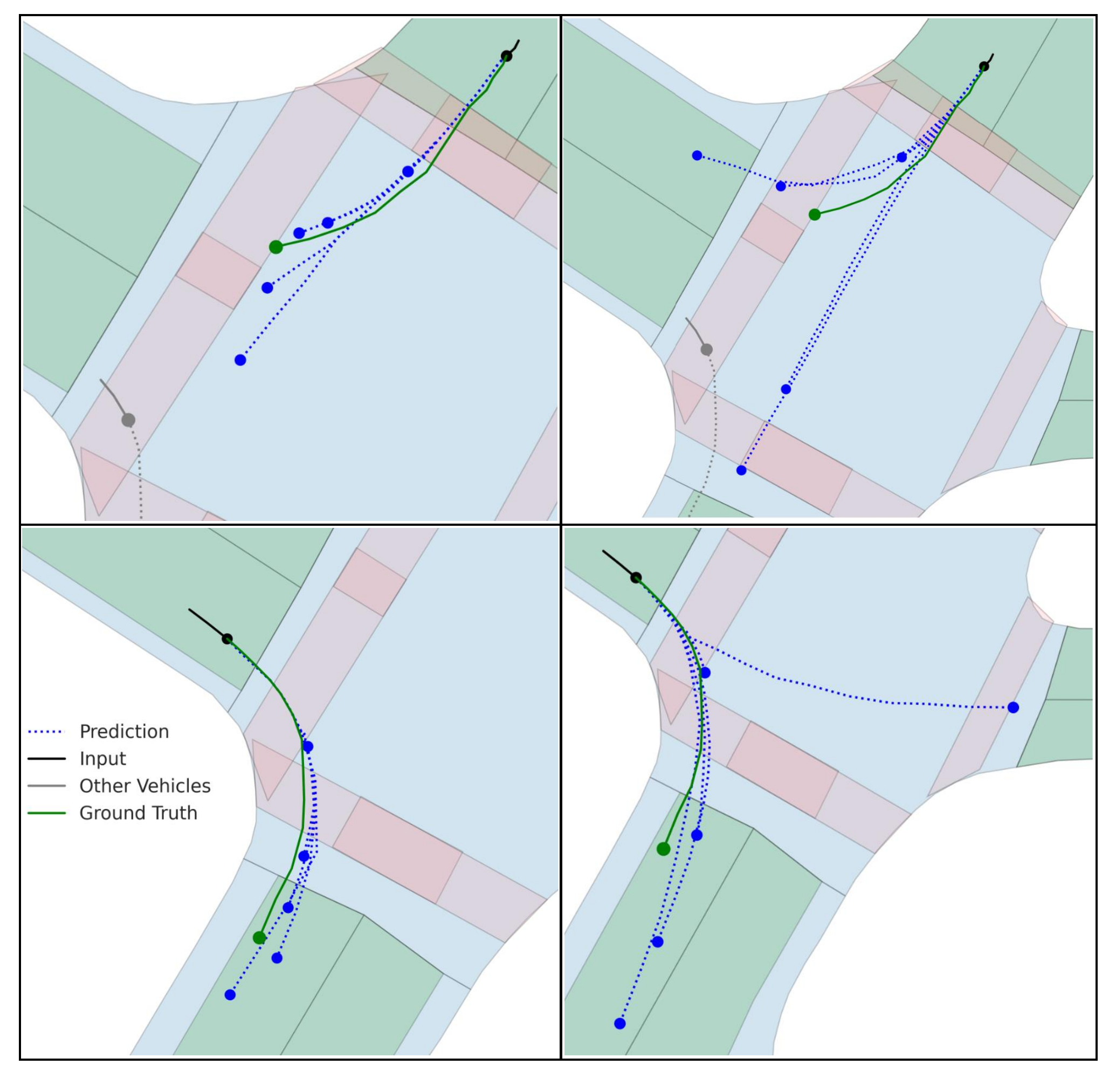}
    \caption{Left column shows the predicted trajectories by the network without mode queries. The right column shows the corresponding scenarios after the mode queries are incorporated into the network. The generalization capability of the network improves with mode queries as the network can predict the trajectories that can follow multiple scene-consistent paths.}
    \label{fig:mode_collapse}
\end{figure}

The recurrent decoder employs deformable attention \cite{zhu2021deformable} to gather essential information from the scene encodings. The deformable attention module consists of deformable self-attention and deformable cross-attention modules. Thereby, the scene encodings are first encoded in the deformable self-attention module, which performs multi-scale feature fusion. The position information in the scene encodings is captured with non-learnable sinusoidal positional embeddings \cite{zhu2021deformable}. The fused scene encodings are then processed by a deformable cross-attention module, in which the attention map is learned through a linear transformation of queries. During our initial experiments, we only introduced temporal queries corresponding to each mode. The objective of the temporal queries was two-fold, first, they must learn the temporal correlation across the different time steps in the predicted trajectories, and second, they must distinguish between different modes as illustrated in previous works \cite{girgis2021latent,varadarajan2022multipath++,zhou2023query}. However, our preliminary experiments showed that this setup results in mode collapse (see the left column of Figure \ref{fig:mode_collapse}). We observed that although the different modes do correspond to different speeds, they miss out on other possible scene-consistent trajectories. To overcome this issue, we use another set of queries, called mode queries, in our network architecture. The results show that mode queries significantly improve the diversity of modes (see the right column of Figure \ref{fig:mode_collapse}). Another aspect of the original deformable cross-attention \cite{zhu2021deformable} is that it utilizes reference points to help the network focus its attention at a particular location in the image. We exploit this property of deformable cross-attention and set the reference point to the ego vehicle position based on the recurrent predicted trajectory output. 

\begin{figure}[t]
        \centering
	\includegraphics[width=6.5cm]{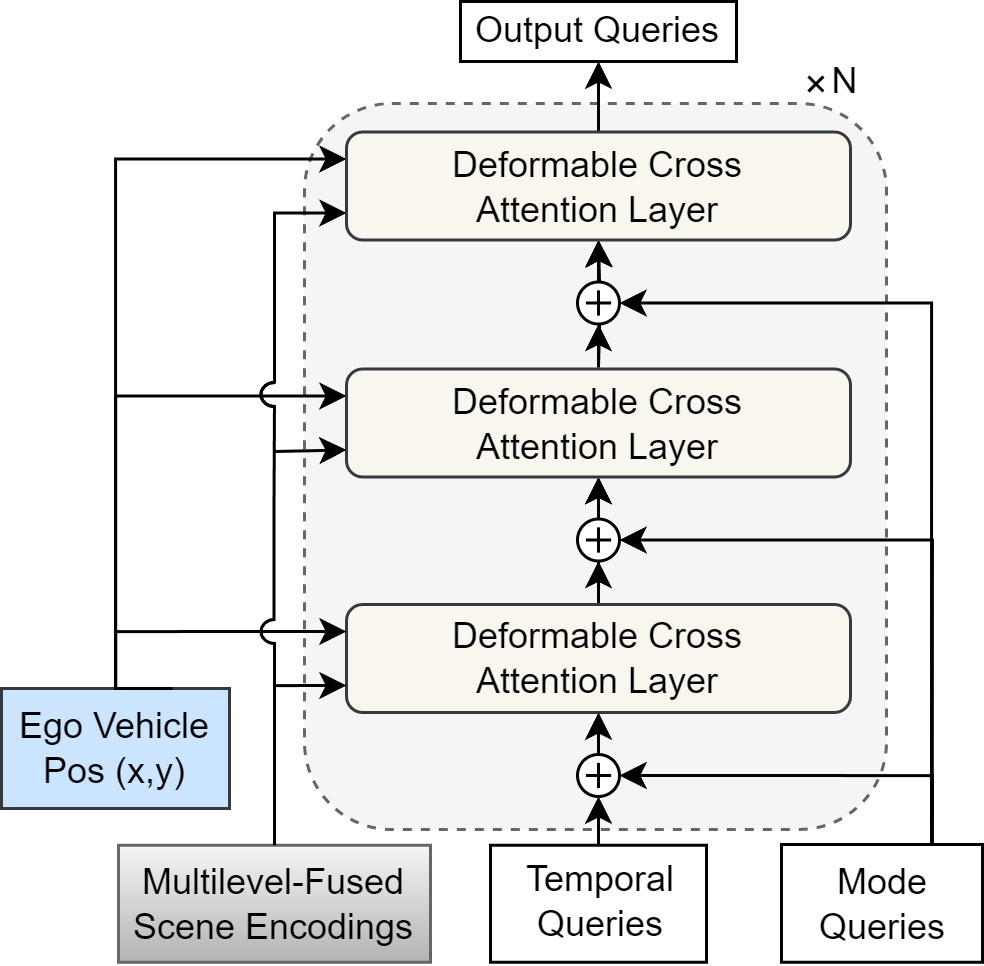}
	\caption{An illustration of the proposed deformable cross-attention module. The offsets in the deformable cross-attention layer are computed with the linear transformation of the queries (as is done in original deformable attention \cite{zhu2021deformable}). These queries are generated by summing up temporal queries and mode queries. Values are then sampled from the multi-scale fused scene encodings at these offset locations and a weighted sum of the sampled values is computed. This process is repeated N times to produce the output queries.}
 \label{fig:multi-layer-decoder}
\end{figure}

The recurrent behavior in the decoder is achieved by incorporating a feedback loop into the deformable cross-attention module. It outputs queries corresponding to individual modes, which are then transformed into multi-modal trajectories using \ac{mlp}. To capture the temporal correlation in the predicted trajectories, the temporal queries are updated to output queries of the previous iteration. In addition, the reference point is updated to the end point of the predicted trajectories from the previous iteration. 

The working mechanism of the deformable cross-attention module is shown in Figure \ref{fig:multi-layer-decoder}. It consists of multiple iterations of deformable cross-attention layers between queries and fused scene encodings. The mode queries are added to the temporal queries before every deformable cross-attention layer. 



\subsection{Loss Formulation}
\label{section:loss_formulation}
We use the loss function proposed by HiVT \cite{zhou2022hivt}. It encourages diversity in predicted trajectories by optimizing only the best mode. The selection of the best mode is done based on the minimum $l_2$ between the ground truth and the predicted trajectories, averaged over all time steps. The loss function comprises of a regression loss $\mathcal{L}_{reg}$ and a classification loss $\mathcal{L}_{cls}$:

\begin{equation}
    \mathcal{L} = \mathcal{L}_{reg} + \mathcal{L}_{cls},
\end{equation}

Regression loss optimizes negative log-likelihood with the probability density function of the Laplace distribution, $\mathbb{L(\cdot \mid \cdot)}$, as follows:

\begin{equation}
    \mathcal{L}_{reg} = -\frac{1}{T_o} \sum_{t=1}^{T_o} log[\mathbb{L}(P_t \mid \mu_t, b_t)], 
\end{equation}

\noindent where $\mu_t$, and $b_t$ are the position and uncertainty at each time step of the predicted best mode trajectory respectively, and $P_t$ are the ground truth trajectory positions. The classification loss aims to optimize only the mode probabilities $\pi(k)$ corresponding to mode $k$ using the cross-entropy loss:

\begin{equation}
    \mathcal{L}_{cls} = - \frac{1}{M} \sum_{k=1}^{M}  log(\pi(k))  \mathbb{L}(P_{T_o, k} \mid \mu_{T_o, k}, b_{T_o, k}),
\end{equation}

\section{Experiments}\label{section:experiments}
This section focuses on the experiments conducted using \ac{caspformer}. Section \ref{subsection:experimental_setup} illustrates the dataset, metrics, and other experimental setting. Section \ref{subsection:results} provides a detailed comparison with the current state-of-the-art. Section \ref{subsection:ablation_studies} explains the design context of the ablation studies and the corresponding results. 

\subsection{Experimental Setup}\label{subsection:experimental_setup}

\noindent \textbf{Dataset.} We test \ac{caspformer} on the publicly available nuScenes dataset \cite{caesar2020nuscenes}, which contains 1000 twenty-second-long traffic scenes from Boston and Singapore. The dataset consists of various traffic situations. 

\noindent \textbf{Metrics.} We report the performance of \ac{caspformer} using minADE\textsubscript{k}, MR\textsubscript{k}, minFDE\textsubscript{k}, and OffRoadRate. minADE\textsubscript{k} computes the average of pointwise $l_2$ distance in meters between the ground truth and the predicted modes and then chooses the minimum value across all $k$ modes. minFDE\textsubscript{k} computes the $l_2$ distance between the ground truth and predicted modes for the last time step only, and then selects the minimum amongst all $k$ modes. MR\textsubscript{k} is defined as the fraction of misses, where a miss occurs if the maximum pointwise $l_2$ distance between the ground truth and the predicted modes is more than two meters. OffRoadRate measures the fraction of predicted trajectories that lie outside the driving area.  

\noindent \textbf{Implementation Details.} \ac{caspformer} is trained on an Nvidia A100 GPU with a batch size of 64 using AdamW optimizer \cite{loshchilov2018decoupled}. The static and dynamic contexts cover a region of size 152 m $\times$ 96 m with a resolution of 1 m, leading to the input grid sizes of (152, 96). The ego vehicle is placed at (122, 48) pointing upward in this grid. We perform data augmentation on the rasterized inputs during training. The inputs are randomly rotated in between $[-60^{\circ}, 60^{\circ}]$, and randomly translated in between $[-3, 3]$ with a probability of 0.75. The number of past time steps for dynamic context is set to $T_i=3$, which is equivalent to 1 s of input trajectory as the sampling rate is 2 Hz. The number of future time steps for the output is set to $T_o=12$, which is equivalent to 6 s of prediction. The number of modes is set to $M=5$. The value of repetitions of deformable attention layers $N$, as depicted in  Figure \ref{fig:multi-layer-decoder}, is set to four. The number of feature levels in the feature pyramid is also set to four, and the hidden dimension of all feature maps is set to 64. 

\begin{table*}[!b]
    \centering
    \caption{Comparison with state-of-the-art on the nuScenes prediction test split.}
    \label{tab:experiment_table}
    \begin{tabular}{c|c c|c|c}
    \hline
    Method & minADE\textsubscript{5}$\downarrow$ & MR\textsubscript{5}$\downarrow$ & minFDE\textsubscript{1}$\downarrow$ & OffRoadRate$\downarrow$ \\
    \hline
    GOHOME \cite{gilles2022gohome}  & 1.42 & 0.57 & 6.99 & 0.04 \\
    Autobot \cite{girgis2021latent}  & 1.37 & 0.62 & 8.19 & 0.02 \\
    THOMAS \cite{gilles2022thomas}  & 1.33 & 0.55 & 6.71 & 0.03 \\
    PGP \cite{deo2021multimodal}  & 1.27 & 0.52 & 7.17 & 0.03 \\
    MacFormer \cite{feng2023macformer}  & 1.21 & 0.57 & 7.50 & 0.02 \\
    LAFormer \cite{liu2023laformer}  & 1.19 & \textbf{0.48} & 6.95 & 0.02 \\
    FRM \cite{park2023leveraging}  & 1.18 & \textbf{0.48} & 6.59 & 0.02 \\
    Q-EANet\_v2 \cite{chen2023q} & 1.18 & \textbf{0.48} & 6.77 & 0.03 \\
    CASPNet\_v2 \cite{schafer2023caspnet++} & 1.16 & 0.50  & \textbf{6.18} & \textbf{0.01} \\
    \hline
    CASPFormer (ours) & \textbf{1.15} & \textbf{0.48} & 6.70 & \textbf{0.01} \\
    \hline
    \end{tabular}
\end{table*}

\subsection{Results}\label{subsection:results}
\begin{figure}[t]
    \centering
    \includegraphics[width=122mm]{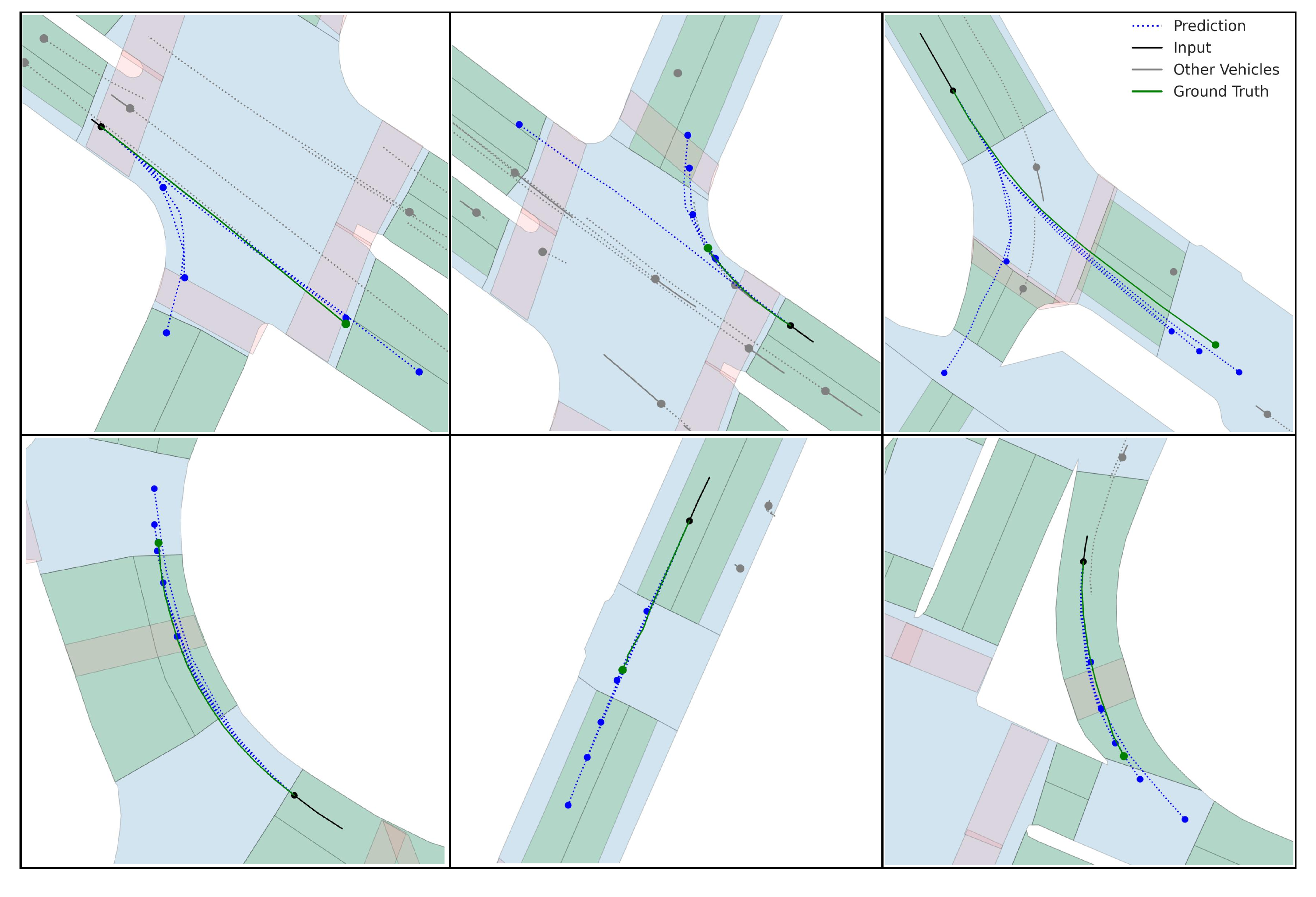}
    \caption{Qualitative results on nuScenes prediction validation split. The blue region represents the drivable area and the green overlays portray the driving lanes. The pedestrian crossing is shown in red color. The current position of the ego vehicle is indicated with the black dot at the end of the input trajectory. The network can predict multiple scene-consistent trajectories in diverse scenarios such as intersections and crossings.}
    \label{fig:qulaititative_results}
\end{figure}

We compare our work against the state-of-the-art on the nuScenes Motion Prediction Challenge \cite{nuscene2020prediction} in Table  \ref{tab:experiment_table}. \ac{caspformer} achieves the best performance in minADE\textsubscript{5}, MR\textsubscript{5}, and OffRoadRate. It should be noted that we have not included the work by Yao et al. \cite{yao2023goal} in our comparison, as their model Goal-LBP performs significantly worse on minFDE\textsubscript{1} (9.20) and OffRoadRate (0.07) in comparison to all other methods mentioned in Table \ref{tab:experiment_table}. Moreover, this study is published after the conclusion of our work and therefore its methods could not have been verified and considered in our approach. Our qualitative results are illustrated in Figure \ref{fig:qulaititative_results}, which shows that \ac{caspformer} can predict multiple modes consistent with the scene. In addition, we discover that each mode corresponds to a different driving speed of the ego vehicle. A potential limitation is that in some cases the trajectories are not well aligned with the lanes and we aim to tackle this in our future work.

\subsection{Ablation Studies}\label{subsection:ablation_studies} 
We perform ablation studies on the nuScenes prediction validation split. The results of our ablation study are shown in Table \ref{tab:ablation}. Where experiment $\#1$ represents the baseline network architecture, which includes all modules, as presented in Figure \ref{fig:recurrent-decoder}. In the following, we discuss the experimental setting of all the ablation studies and their results:

\begin{table*}[t]
    \centering
    \caption{Ablation Study on nuScenes Prediction Validation Split}
    \label{tab:ablation}
    \resizebox{\columnwidth}{!}{
    \begin{tabular}{c|c c c c|c c|c}
    \hline
    \# & \makecell{Mode \\ Embeddings} & \makecell{Deformable \\ Self Attention} & \makecell{Recurrent \\ Architecture} & \makecell{Ego Vehicle \\ Position} & minADE\textsubscript{5}$\downarrow$ & MR\textsubscript{5}$\downarrow$ & minFDE\textsubscript{1}$\downarrow$ \\
    \hline
    1. &\checkmark & \checkmark & \checkmark & \checkmark & \textbf{1.13} & \textbf{0.46} & \textbf{6.43} \\
    2. & - & \checkmark & \checkmark & \checkmark & 1.72 & 0.60 & 6.60 \\
    3. &\checkmark & - & \checkmark & \checkmark & 1.26 & 0.53 & 6.92 \\
    4. &\checkmark & \checkmark & - & \checkmark & 1.21 & 0.48 & 6.63 \\
    5. &\checkmark & \checkmark & \checkmark & - & 1.15 & 0.48 & 6.51 \\
    \hline
    \end{tabular}
    }
\end{table*}

\noindent \textbf{Importance of mode queries.} To show the significance of mode queries, we conduct an experiment, in which the mode queries are not provided as input to deformable cross-attention module, as presented in Figure \ref{fig:recurrent-decoder}. The results of this experiment illustrate that the network performs worse on all metrics especially on minADE\textsubscript{5} when the mode queries are not provided in comparision to when they are (see experiments $\#1$ and $\#2$ in Table \ref{tab:ablation}). The corresponding qualitative results of the experiment $\#2$ are illustrated in Figure \ref{fig:mode_collapse}, which indicate that even though the modes retain the property of capturing various speeds of the ego vehicle, they follow the same path and miss out on other possible paths, thus leading to mode collapse. Therefore, we deduce that the introduction of mode queries helps avoid mode collapse in \ac{caspformer}.

\noindent \textbf{Effect of Deformable Self-Attention.} The studies \cite{roh2021sparse,li2023lite} point out that a significant computational cost in deformable attention comes from its deformable self-attention module. In our experiments, we also discover that if the deformable self-attention module is removed, the training time reduces by $60.3\%$, while minADE\textsubscript{5}, MR\textsubscript{5} and minFDE\textsubscript{1} increase by 11.5$\%$, 15.2$\%$ and 7.6$\%$ respectively (see experiments $\#1$ and $\#3$ in Table \ref{tab:ablation}). This can be a reasonable trade-off depending on the constraints for the motion prediction module. When removing the deformable self-attention module, we sum up the positional embeddings and scene encodings along the channel dimension and provide it directly as input into the deformable cross-attention module.


\noindent \textbf{Importance of Recurrent Architecture.} 
We also test whether the recurrent feedback loops help the network in performing better across the various metrics. Thus we remove both feedback loops from our baseline network (as shown in Figure \ref{fig:recurrent-decoder}) and decode the complete 6 s trajectories in a single forward pass. The results of this experiment show that the performance of the network decreases across all the metrics when the feedback loops are not present in the network (see experiments $\#1$ and $\#4$ in Table \ref{tab:ablation}). This confirms the findings of the works \cite{schafer2022context,zhou2023query} that the recurrent architecture improves multimodal trajectory prediction. 

\noindent \textbf{Importance of Providing Ego Vehicle Position.}
The results of our experiments show that setting the reference point to the ego vehicle position does not improve the network performance by any significant degree (see experiments $\#1$ and $\#5$ in Table \ref{tab:ablation}), where in the experiment $\#5$, the reference points are directly learned via linear transformation of mode embeddings as is the case with the original deformable attention \cite{zhu2021deformable}. Nevertheless, we speculate that setting the reference point to the position of the agent in the scene can play an important role in multi-agent joint motion prediction, and leave a detailed study of this for future work.




\section{Conclusion}\label{section:conclusion}
In this study, a novel network architecture, \ac{caspformer}, is proposed which performs multi-modal trajectory prediction from \ac{bev} images of the surrounding scene. \ac{caspformer} employs a deformable attention mechanism to decode trajectories recurrently. Moreover, our work illustrates a mechanism to incorporate mode queries, which prevents the mode collapse and enables the network to generate scene-consistent multi-modal trajectories. We also identify that excluding the deformable self-attention module leads to a significant decrease in computational cost, without much effect on the network performance. Thus, in our future work, we aim to remove or modify the deformable self-attention module. Moreover, our future work would involve further study of the effect of vectorized dynamic context and the impact of reference points in multi-agent joint motion prediction.










\bibliography{root}{}%
\bibliographystyle{splncs04}

\end{document}